\documentclass[letterpaper, 10 pt, conference]{ieeeconf}  

\IEEEoverridecommandlockouts                              

\usepackage{graphics} 
\usepackage{epsfig} 
\usepackage{amsmath} 
\usepackage{amssymb} 

\usepackage{booktabs}
\usepackage{multirow}
\usepackage{subcaption}
\usepackage{xcolor}
\usepackage{graphicx}

\title{\LARGE \bf
Learning Reliable Parking Policies via Offline Reinforcement Learning with Quantized Action Representations 
}

\author{Zewei Yang, Zengqi Peng, and Jun Ma, \textit{Senior Member, IEEE}
\thanks{Zewei Yang and Zengqi Peng are with the Robotics and Autonomous Systems Thrust, The Hong Kong University of Science and Technology (Guangzhou), Guangzhou 511453, China (e-mail: zyang363@connect.hkust-gz.edu.cn; zpeng940@connect.hkust-gz.edu.cn). }
\thanks{Jun Ma is with the Robotics and Autonomous Systems Thrust, The Hong Kong University of Science and Technology (Guangzhou), Guangzhou 511453, China, and also with the Cheng Kar-Shun Robotics Institute, The Hong Kong University of Science and Technology, Hong Kong SAR, China (e-mail: jun.ma@ust.hk).} 
}

\begin{document}

\maketitle
\thispagestyle{empty}
\pagestyle{empty}

\begin{abstract}
Parking is a routine yet safety-critical task for autonomous vehicles operating in urban environments. However, cluttered and weakly structured parking spaces, compounded by the interactive uncertainty from surrounding vehicles, hinder reliable maneuver generation. To address these challenges, we develop a waypoint-level offline reinforcement learning framework for interaction-aware autonomous parking. Specifically, a dedicated parking dataset is constructed from hierarchical expert rollouts with rotational waypoint augmentation, covering both non-interactive scenarios and interactive ones. The policy is then conditioned on a compact state representation, in which LiDAR-based obstacle features are adapted to the target pose via feature-wise linear modulation. A state-conditioned tokenizer further quantizes continuous waypoint sequences into discrete action tokens, over which conservative Q-learning is performed to suppress value overestimation on poorly supported actions. Extensive closed-loop experiments are conducted in the high-fidelity CARLA simulator. The proposed framework attains the highest parking success rate among all baselines and transfers reliably to unseen parking slots.
\end{abstract}

\section{INTRODUCTION}

Autonomous parking is an essential capability for intelligent vehicles, yet it remains challenging in complex parking environments \cite{li2022online}. Unlike structured road driving, parking is often performed in narrow, cluttered, and weakly structured spaces, where lane-level guidance and environmental regularity are limited \cite{li2021optimization}. These scenarios provide only small maneuvering margins and require the ego vehicle (EV) to reason about surrounding vehicles, static obstacles, and other moving agents \cite{leu2022autonomous}. Autonomous parking therefore requires maneuver generation that is safe, efficient, and robust under tight spatial constraints and interactive uncertainty.

Traditional methods commonly formulate autonomous parking as a constrained trajectory-optimization problem, in which vehicle kinematics, collision avoidance, environmental constraints, and maneuver feasibility are explicitly modeled \cite{li2021optimization}. Such model-based formulations are interpretable and allow safety and feasibility requirements to be directly encoded. However, they often rely on accurate scene modeling and extensive expert design, including hand-crafted objectives, constraints, and planning heuristics \cite{huang2024dtpp}. Their efficiency and robustness may also degrade when nonlinear optimization faces complex constraints, poor initialization, or frequent replanning, especially in narrow, cluttered, or time-varying parking scenes \cite{lian2023trajectory}.

To reduce the reliance on hand-engineered planning logic, imitation learning (IL) has been explored for learning parking maneuvers from expert demonstrations or optimized trajectories. Once trained, IL-based planners \cite{li2024parkinge2e} can efficiently generate maneuver trajectories directly from raw sensory inputs. However, their closed-loop performance depends strongly on the coverage and quality of the training data. When the vehicle visits states that are poorly represented in the demonstrations, prediction errors may accumulate over time and lead to unstable or suboptimal maneuvers \cite{lu2023imitation}.

Reinforcement learning (RL) provides a complementary reward-driven formulation for sequential decision making \cite{mnih2015human}. In autonomous parking, RL can support long-horizon maneuver generation while jointly considering safety, efficiency, and parking quality \cite{tang2023path}. Online RL, however, is difficult to apply to parking tasks because trial-and-error interaction is costly, and exploratory behavior in confined spaces may be unsafe. Training is further complicated by delayed rewards and continuous or high-dimensional action spaces \cite{han2022off}. These challenges motivate offline RL, which learns from fixed datasets while retaining the potential for reward-guided policy improvement without additional online interaction.

\begin{figure*}[t]
  \centering
  \includegraphics[width=\textwidth]{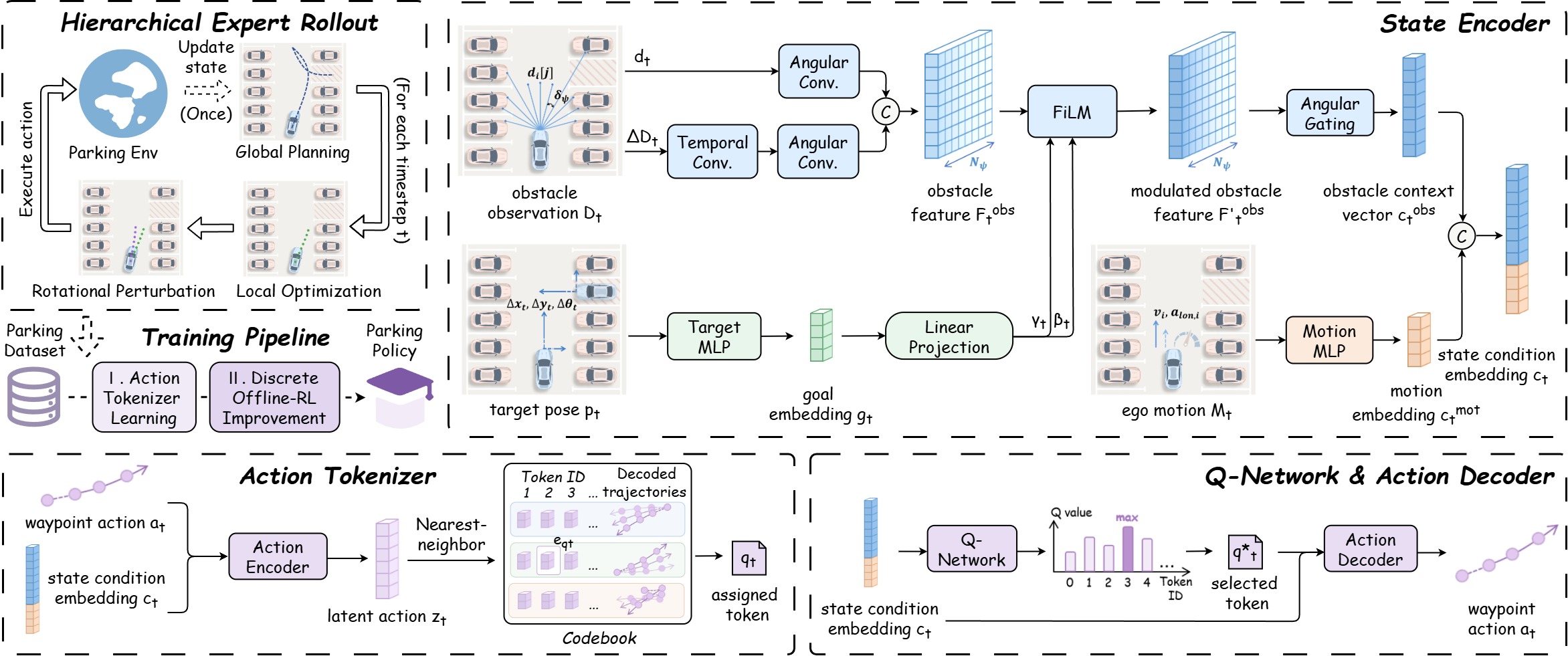}
  \caption{Overall pipeline of the proposed waypoint-level offline RL framework for autonomous parking.}
  \label{fig:framework}
\end{figure*}

In this work, we propose an interaction-aware offline RL framework that generates parking maneuvers as waypoint sequences rather than low-level control commands. Fig.~\ref{fig:framework} illustrates the overall pipeline of the proposed method. We construct a planning-oriented offline dataset from hierarchical expert rollouts with rotational waypoint augmentation, spanning both non-interactive and interactive parking scenarios. The state encoder extracts compact obstacle representations from LiDAR histories and integrates target-pose information through feature-wise linear modulation (FiLM). To facilitate offline policy learning, we introduce a state-conditioned action tokenizer that quantizes continuous waypoint sequences into discrete action tokens. A conservative value-learning objective is then applied in the discrete action space to suppress overestimation on poorly supported tokens and select reward-favorable maneuvers from the offline data. Our main contributions are summarized as follows:

\begin{itemize}
    \item A waypoint-level offline RL framework is developed for autonomous parking tasks. It discretizes continuous waypoint sequences into state-conditioned action tokens and performs conservative value learning, yielding reward-guided policies without any online interaction.
    \item A task-specific parking dataset is constructed from hierarchical expert rollouts with rotational waypoint augmentation. It spans both non-interactive cases and interactive ones in which the EV must coordinate with surrounding traffic.
    \item The proposed framework is extensively evaluated through closed-loop experiments in the CARLA simulator. It achieves higher parking success, lower collision risk, and reliable generalization to unseen slots compared with representative baselines.
\end{itemize}

\section{RELATED WORK}

\subsection{Model-Based Planning for Autonomous Parking}

Model-based parking planning remains a widely adopted paradigm, as it can explicitly incorporate feasibility and safety constraints. Existing studies have improved this paradigm from several perspectives. One line of work reduces the computational burden of collision checking and constraint handling through lightweight iterative optimization \cite{li2021optimization}. Another line improves initialization by using enhanced Hybrid A* to generate feasible seeds for nonlinear refinement \cite{lian2023trajectory}, or by optimizing only critical variables to obtain lower-dimensional initial guesses \cite{guo2022down}. For time-varying parking scenes, online replanning methods update only the affected trajectory segment rather than resolving the entire maneuver from scratch \cite{li2022online2}. These methods improve feasibility, computational efficiency, and replanning responsiveness in their respective settings. Nevertheless, they still depend on accurate scene models, expert-designed formulations, and reliable optimization under complex parking conditions.

\subsection{Imitation Learning for Autonomous Parking}

IL-based parking planners reduce the need for online optimization by learning parking behaviors from expert trajectory data. Early studies often embedded learned components into conventional planning pipelines. For example, learned feasible-trajectory distributions can bias Hybrid A* expansion to improve search efficiency \cite{kim2023neural}, while recurrent deep neural networks can approximate optimal parking trajectories from vehicle kinematic states \cite{chai2022deep}. More recent end-to-end methods shift the focus toward perception-driven planning. ParkingE2E \cite{li2024parkinge2e} predicts future waypoints from camera observations and target information, whereas MultiPark \cite{zheng2025multipark} adopts a transformer-based formulation to generate diverse feasible trajectory segments for multimodal parking maneuvers. Overall, these methods improve online efficiency and reduce manual planner design. However, their closed-loop reliability remains constrained by the diversity and coverage of the training data, particularly when rare configurations or recovery behaviors are insufficiently represented.

\subsection{Reinforcement Learning for Autonomous Parking}

RL-based parking planning learns policies from task rewards rather than supervised trajectory targets. Improved Monte Carlo tree search has been used for time-efficient parallel parking by introducing reward-based search into maneuver generation \cite{song2022time}. Model-free actor--critic methods further learn parking and tracking policies under non-ideal conditions \cite{tang2023path}. Other studies incorporate structural priors into RL-based planning, including hybrid policies guided by geometric planning cues \cite{jiang2025hope} and hierarchical formulations with kinematic constraints \cite{xu2025hrl}. These methods support long-horizon decision making and policy adaptation through reward feedback. However, they typically require extensive interaction, careful reward design, and exploratory trials, which can be costly or unsafe in confined parking environments. This limitation motivates offline RL as a way to retain reward-guided policy improvement while avoiding additional trial-and-error interaction.

\section{PRELIMINARIES}

In this work, autonomous parking is formulated as a discounted Markov decision process (MDP),
\begin{equation}
\mathcal{M} = (\mathcal{S}, \mathcal{A}, \mathcal{P}, r, \gamma),
\end{equation}
where $\mathcal{S}$, $\mathcal{A}$, $\mathcal{P}$, $r$, and $\gamma$ denote the state space, action space, transition kernel, reward function, and discount factor, respectively.

\textbf{State Space $\mathcal{S}$:}
At decision step $t$, the state $s_t\in\mathcal{S}$ is defined as
\begin{equation}
s_t = (D_t, M_t, p_t),
\end{equation}
where $D_t$ encodes recent obstacle observations, $M_t$ describes recent ego-motion history, and $p_t$ represents the pose error relative to the target parking pose. To provide short-term perception and motion context, $D_t$ and $M_t$ are constructed from the most recent $N_h$ time steps. Specifically,
\begin{equation}
D_t = (d_{t-N_h+1}, \ldots, d_t),
\end{equation}
where each $d_i\in\mathbb{R}^{N_{\psi}}$ is an obstacle-distance vector obtained from LiDAR-based perception over $N_{\psi}$ sampled rays. The $j$-th entry of $d_i$ records the distance to the nearest obstacle along the ray at angle $\psi_j=(j-1)\delta_{\psi}$ in the ego frame, where $\delta_{\psi}$ denotes the angular resolution. Similarly,
\begin{equation}
M_t = (m_{t-N_h+1}, \ldots, m_t),
\end{equation}
where $m_i=[v_i,\,a_{\mathrm{lon},i}]^\top$ contains the ego speed and longitudinal acceleration at time step $i$. The target-pose error is defined as
\begin{equation}
p_t = [\Delta x_t,\, \Delta y_t,\, \Delta \theta_t]^\top,
\end{equation}
where $\Delta x_t$, $\Delta y_t$, and $\Delta \theta_t$ denote the relative position and heading errors with respect to the target parking pose.

\textbf{Action Space $\mathcal{A}$:}
The action $a_t\in\mathcal{A}$ is represented as an ordered sequence of future waypoints in the ego frame, with the rear-axle center as the origin:
\begin{equation}
a_t = (w_{t,1}, \ldots, w_{t,H}),
\end{equation}
where $H$ is the action horizon and each waypoint
\begin{equation}
w_{t,k} = [x_{t,k},\, y_{t,k},\, \theta_{t,k}]^\top
\end{equation}
specifies the relative pose of the $k$-th future waypoint. The waypoint sequence serves as a short-horizon reference trajectory and is tracked by a downstream controller during execution.

\textbf{State Transition Dynamics:}
The transition kernel $\mathcal{P}(s_{t+1}\mid s_t,a_t)$ characterizes the environment transition induced by executing action $a_t$ at state $s_t$. The transition model is not assumed to be available in closed form.

\textbf{Reward Function:}
The reward function encourages successful and collision-free parking while accounting for maneuver efficiency and control smoothness. Its detailed design is presented in Section~\ref{sec:reward_design}.

\textbf{Discount Factor:}
The discount factor $\gamma\in(0,1)$ weights future rewards in the cumulative return.

\section{METHODOLOGY}

\subsection{Hierarchical Offline Dataset Generation}
\label{sec:dataset_collection}

In offline RL, policy optimization is performed on a fixed dataset without additional environment interaction during training. We construct the offline dataset for autonomous parking using a hierarchical planning-and-control pipeline. The EV is modeled by the following kinematic bicycle model:
\begin{equation}
\begin{bmatrix}
\dot{x}(\tau)\\
\dot{y}(\tau)\\
\dot{\theta}(\tau)\\
\dot{v}(\tau)
\end{bmatrix}
=
\begin{bmatrix}
v(\tau)\cos\theta(\tau)\\
v(\tau)\sin\theta(\tau)\\
v(\tau)\tan\delta(\tau) / L_w\\
a_{\mathrm{lon}}(\tau)
\end{bmatrix},
\label{eq:vehicle_dynamics}
\end{equation}
where $\xi(\tau)=[x(\tau),\,y(\tau),\,\theta(\tau),\,v(\tau)]^\top$ denotes the vehicle state, comprising the planar position $x(\tau)$ and $y(\tau)$, heading angle $\theta(\tau)$, and longitudinal speed $v(\tau)$; $u(\tau)=[\delta(\tau),\,a_{\mathrm{lon}}(\tau)]^\top$ denotes the control input consisting of the steering angle $\delta(\tau)$ and longitudinal acceleration $a_{\mathrm{lon}}(\tau)$; and $L_w$ is the wheelbase.

\subsubsection{Global Reference Generation}

Given the initial vehicle pose and the target parking pose, Hybrid A* first generates a coarse geometric path. A velocity profile is then assigned along this path using a discretized form of the vehicle dynamics in Eq.~\eqref{eq:vehicle_dynamics}, yielding a dynamically feasible global reference trajectory.

\subsubsection{Local Trajectory Optimization}

At each planning step $t$, a fixed-horizon local reference sequence $\boldsymbol{\bar{\xi}}_{t}=(\bar{\xi}_{t,1},\ldots,\bar{\xi}_{t,H})$ is extracted from the global reference trajectory. The sequence starts from a local projection of the current vehicle pose onto the global reference. Given $\boldsymbol{\bar{\xi}}_{t}$, a nonlinear programming (NLP) problem is solved to obtain an optimized local state sequence $\boldsymbol{\xi}^{*}_{t}=(\xi^{*}_{t,1},\ldots,\xi^{*}_{t,H})$:
\begin{equation}
\min_{\boldsymbol{\xi}_{t},\,\boldsymbol{u}_{t}}
\sum_{k=1}^{H}
\left\| \xi_{t,k}-\bar{\xi}_{t,k} \right\|_{W_{\xi}}^{2}
+
\sum_{k=1}^{H-1}
\left\| u_{t,k+1}-u_{t,k} \right\|_{W_u}^{2},
\end{equation}
where $W_{\xi}$ and $W_u$ are weighting matrices for reference tracking and control variation, respectively. The NLP is discretized using a forward-Euler form of Eq.~\eqref{eq:vehicle_dynamics}, with actuation limits and obstacle-clearance constraints enforced over the planning horizon. The optimized waypoint action $a_t^*=(w_{t,1}^*,\ldots,w_{t,H}^*)$ is extracted from the pose components of $\boldsymbol{\xi}_{t}^{*}$, transformed into the ego frame, and treated as the expert action for the current state $s_t$.

\subsubsection{Waypoint Action Perturbation}

To improve maneuver diversity in the offline dataset, we apply rotational perturbations to the expert action \cite{cai2025navdp}. Given the expert waypoint action $a_t^*$, a perturbation angle $\alpha_t$ is sampled, and a perturbed waypoint action $a_t$ is generated by rotating each waypoint in the ego frame:
\begin{equation}
\begin{bmatrix}
x_{t,k}\\
y_{t,k}
\end{bmatrix}
=
R(\alpha_t)
\begin{bmatrix}
x_{t,k}^*\\
y_{t,k}^*
\end{bmatrix},
\quad
\theta_{t,k}
=
\theta_{t,k}^*+\alpha_t,
\end{equation}
where $R(\alpha_t)\in\mathbb{R}^{2\times2}$ denotes the planar rotation matrix. This perturbation increases local action diversity around expert behaviors while preserving the geometric structure of the waypoint sequence, exposing policy learning to a broader distribution of short-horizon parking maneuvers.

\subsubsection{Transition Collection}

During data collection, the perturbed waypoint action $a_t$ is tracked by a linear quadratic regulator (LQR). After the corresponding control command is applied for one decision interval, the vehicle state is updated through sensing and localization, and the resulting transition is recorded with its reward. The offline dataset is organized as
\begin{equation}
\mathcal{D}
=
\{(s_t,a_t,r_t,s_{t+1},\eta_t)\}_{t},
\end{equation}
where $\eta_t\in\{0,1\}$ is a terminal indicator specifying whether the parking episode terminates.

\subsection{Network Architecture}

The network follows a modular design that separates state representation, action discretization, and token-level value estimation. The state encoder produces a compact condition embedding from the structured parking state. The tokenizer maps continuous waypoint actions to discrete action tokens, while the Q-network evaluates these tokens for action selection.

\subsubsection{State Encoder}

For the obstacle-distance history $D_t\in\mathbb{R}^{B\times N_h\times N_\psi}$, the obstacle branch captures both the current obstacle layout and short-term distance variations. It processes the current distance observation $d_t$ and temporal distance differences derived from $D_t$ through separate streams. The temporal-difference stream uses a per-angle temporal convolution to model distance changes along each ray. The two streams are further refined in parallel by angular residual convolutions with circular padding \cite{hu2024rangeldm}, which aggregate neighboring rays while preserving the circular range-view topology. The refined stream features are then fused to obtain the obstacle feature map $F_t^{\mathrm{obs}}\in\mathbb{R}^{B\times C_{\mathrm{obs}}\times N_\psi}$.

The target-pose error $p_t\in\mathbb{R}^{B\times 3}$ is encoded by a multilayer perceptron (MLP) into a goal embedding $g_t\in\mathbb{R}^{B\times C_{\mathrm{goal}}}$. To inject target-pose information into the obstacle representation, FiLM \cite{perez2018film} coefficients $\gamma_t,\beta_t\in\mathbb{R}^{B\times C_{\mathrm{obs}}\times 1}$ are generated from $g_t$ and applied to the obstacle feature map in a channel-wise manner:
\begin{equation}
\tilde{F}_{t}^{\mathrm{obs}}
=
(1+\gamma_t)\odot F_{t}^{\mathrm{obs}}
+
\beta_t,
\end{equation}
where $\tilde{F}_{t}^{\mathrm{obs}}\in\mathbb{R}^{B\times C_{\mathrm{obs}}\times N_\psi}$ denotes the target-modulated obstacle feature map.

A learned angular gating module pools $\tilde{F}_{t}^{\mathrm{obs}}$ into an obstacle context vector $c_t^{\mathrm{obs}}\in\mathbb{R}^{B\times C_{\mathrm{obs}}}$. In parallel, the ego-motion history $M_t\in\mathbb{R}^{B\times N_h\times 2}$ is flattened and encoded by an MLP into a motion embedding $c_t^{\mathrm{mot}}\in\mathbb{R}^{B\times C_{\mathrm{mot}}}$. The final condition embedding is
\begin{equation}
c_t=[c_t^{\mathrm{obs}},\,c_t^{\mathrm{mot}}]\in\mathbb{R}^{B\times (C_{\mathrm{obs}}+C_{\mathrm{mot}})}.
\end{equation}
Here, $C_{\mathrm{obs}}$, $C_{\mathrm{goal}}$, and $C_{\mathrm{mot}}$ denote the obstacle-feature, goal-embedding, and motion-embedding dimensions, respectively.

\subsubsection{Action Tokenizer}

Rather than directly modeling the high-dimensional continuous action space induced by horizon-level waypoint sequences, we introduce an action tokenizer to learn a discrete token representation for each waypoint action. Inspired by state-conditioned action quantization (SAQ) \cite{luo2023action}, the tokenizer uses the condition embedding to produce context-relevant action tokens, facilitating subsequent offline policy learning.

Given a waypoint action $a_t$ and the condition embedding $c_t$, the action encoder $E_{\psi}$ produces a continuous latent action representation $z_t = E_{\psi}(a_t,c_t)$. A learnable codebook $\mathcal{E}=\{e_q\}_{q\in\mathcal{Q}}$ is then used to quantize $z_t$ by nearest-neighbor assignment:
\begin{equation}
q_t
=
\operatorname*{arg\,min}_{q\in\mathcal{Q}}
\left\|
z_t-e_q
\right\|_2^2,
\end{equation}
where $q_t$ denotes the assigned action token and $e_{q_t}$ is the corresponding codebook embedding.

The action decoder $G_{\omega}$ reconstructs the waypoint action from the selected codebook embedding and the condition embedding:
\begin{equation}
\hat{a}_t = G_{\omega}(e_{q_t},c_t),
\end{equation}
where $\hat{a}_t$ denotes the reconstructed waypoint action.

\subsubsection{Q-Network}

The Q-network parameterizes the action-value function over the action-token vocabulary. Given the condition embedding $c_t$, it outputs $Q_\theta(c_t,\cdot)\in\mathbb{R}^{|\mathcal{Q}|}$, where $|\mathcal{Q}|$ denotes the vocabulary size and the entry indexed by $q$ corresponds to $Q_\theta(c_t,q)$. During execution, the action token is selected greedily as
\begin{equation}
q_t^*=\operatorname*{arg\,max}_{q\in\mathcal{Q}}Q_\theta(c_t,q),
\end{equation}
and decoded by the trained decoder into a continuous waypoint action:
\begin{equation}
a_t=G_{\omega}(e_{q_t^*},c_t).
\end{equation}
The decoded waypoint sequence is then tracked by the downstream LQR controller.

\subsection{Learning Objectives}

The training procedure follows the modular architecture described above. We first learn the tokenizer to obtain action-token assignments for waypoint actions, and then optimize the Q-network on the resulting tokenized transitions with a discrete Conservative Q-Learning (CQL) objective \cite{kumar2020conservative}.

\subsubsection{Action Tokenizer Learning}

The action tokenizer is trained to reconstruct waypoint actions from their assigned codebook embeddings. The tokenizer loss is defined as
\begin{equation}
\mathcal{L}_{\mathrm{tok}}
=
\mathbb{E}_{(s_t,a_t)\sim\mathcal{D}}
\left[
\left\|
a_t-\hat{a}_t
\right\|_2^2
+
\lambda_{\mathrm{com}}
\left\|
z_t-\mathrm{sg}[e_{q_t}]
\right\|_2^2
\right],
\end{equation}
where $\mathrm{sg}[\cdot]$ denotes the stop-gradient operator, and $\lambda_{\mathrm{com}}$ controls the commitment loss. The reconstruction term encourages accurate waypoint recovery, while the commitment term regularizes the continuous latent action representation $z_t$ toward the assigned codebook embedding $e_{q_t}$. The codebook embeddings are updated using an exponential moving average (EMA) scheme.

\begin{table*}[t]
\centering
\caption{Parking performance comparison on the in-distribution slots.}
\label{tab:performance}
\begingroup
\small
\setlength{\tabcolsep}{3.5pt}
\begin{tabular}{ccccccccc}
\toprule
Method & TSR (\%) $\uparrow$ & TFR (\%) $\downarrow$ & CR (\%) $\downarrow$ & TR (\%) $\downarrow$ 
& APE (m) $\downarrow$ & AOE (deg) $\downarrow$ & APT (s) $\downarrow$ & SCT (\%) $\uparrow$ \\
\midrule
SAC     & 0.00  & 3.47  & 34.03 & 62.50 & -- & -- & --    & --    \\
SAC-$N$ ($N=\text{48}$) & 32.64 & 19.44 & 37.50 & 10.42 & 0.70 & 11.56 & 18.73 & 32.64 \\
BC      & 88.19 & 0.00  & 11.81 & 0.00  & 0.74 & 4.75 & 21.54 & 86.85 \\
TD3-BC  & 90.97 & 0.00  & 9.03  & 0.00  & 0.72 & 6.72 & 21.09 & 90.25 \\
Ours    & \textbf{96.53} & 0.69  & \textbf{2.78}  & \textbf{0.00}  & 0.71 & 5.26 & 21.99 & \textbf{95.44} \\
\bottomrule
\end{tabular}
\endgroup
\end{table*}

\subsubsection{Discrete Conservative Q-Learning}

After tokenizer training, each waypoint action $a_t$ is replaced by its assigned action token $q_t$, yielding the tokenized dataset
\begin{equation}
\mathcal{D}_q
=
\{(s_t,q_t,r_t,s_{t+1},\eta_t)\}_t .
\end{equation}
The Q-network is optimized with the following discrete-action CQL objective:
\begin{equation}
\begin{aligned}
\mathcal{L}_{\mathrm{cql}}
&=
\frac{1}{2}\,
\mathbb{E}_{(s_t,q_t,r_t,s_{t+1},\eta_t)\sim\mathcal{D}_q}
\left[
\left(
Q_\theta(c_t,q_t)
-
\mathcal{T}Q_{\bar{\theta}}(c_t,q_t)
\right)^2
\right] \\
&+
\alpha_{\mathrm{cql}}\,
\mathbb{E}_{(s_t,q_t)\sim\mathcal{D}_q}
\left[
\log
\sum_{q\in\mathcal{Q}}
\exp Q_\theta(c_t,q)
-
Q_\theta(c_t,q_t)
\right],
\end{aligned}
\end{equation}
where the Bellman target is defined as
\begin{equation}
\mathcal{T}Q_{\bar{\theta}}(c_t,q_t)
=
r_t
+
\gamma(1-\eta_t)
\max_{q'\in\mathcal{Q}}
Q_{\bar{\theta}}(c_{t+1},q').
\end{equation}
Here, $Q_{\bar{\theta}}$ denotes the target Q-network, and $\alpha_{\mathrm{cql}}$ controls the strength of conservative regularization. The conservative term suppresses overestimated values for candidate action tokens while anchoring the learned value function to tokens observed in the offline dataset.

\subsection{Reward Design}
\label{sec:reward_design}

The reward function encourages accurate, safe, efficient, and smooth autonomous parking behavior. At each decision step, the reward is decomposed as
\begin{equation}
r_t
=
r_{\mathrm{goal}}
+
r_{\mathrm{coll}}
+
r_{\mathrm{len}}
+
r_{\mathrm{ctrl}},
\end{equation}
where $r_{\mathrm{goal}}$ measures parking accuracy, $r_{\mathrm{coll}}$ penalizes collision events, $r_{\mathrm{len}}$ encourages trajectory efficiency, and $r_{\mathrm{ctrl}}$ promotes motion smoothness.

The goal-reaching reward is activated only when the terminal vehicle pose lies within prescribed position and heading thresholds of the target parking pose:
\begin{equation}
r_{\mathrm{goal}}
=
\lambda_{\mathrm{goal}} \,
\mathbb{I}_{\mathrm{reach}} \,
e^{-(\alpha d_{\mathrm{pos}}+\beta d_{\mathrm{head}})},
\end{equation}
where $\lambda_{\mathrm{goal}}$ is the goal-reaching reward weight, $\mathbb{I}_{\mathrm{reach}}$ denotes the reaching indicator, $d_{\mathrm{pos}}=\sqrt{\Delta x_T^2+\Delta y_T^2}$ and $d_{\mathrm{head}}=|\Delta\theta_T|$ denote the terminal position and heading errors, respectively, and $\alpha$ and $\beta$ control the relative sensitivity to position and heading errors.

Safety is encouraged through a collision penalty:
\begin{equation}
r_{\mathrm{coll}}
=
-
\lambda_{\mathrm{coll}} \,
\mathbb{I}_{\mathrm{coll}},
\label{eq:reward_collision}
\end{equation}
where $\lambda_{\mathrm{coll}}$ is the collision penalty weight and $\mathbb{I}_{\mathrm{coll}}$ indicates whether a collision occurs.

Since a finite-horizon waypoint action captures only the local maneuver and does not fully reflect its cost-to-go to the target pose, we construct an NLP-based trajectory continuation initialized from the endpoint of the local trajectory to estimate task-aware efficiency and smoothness. The local trajectory and its continuation are concatenated into a complete candidate trajectory,
\begin{equation}
\{(\xi_i,u_i)\}_{i=1}^{N}
\equiv
\left\{
\left(
x_i,\,
y_i,\,
\theta_i,\,
v_i,\,
\delta_i,\,
a_{\mathrm{lon},i}
\right)
\right\}_{i=1}^{N}.
\end{equation}
The average path-length metric is computed as
\begin{equation}
\ell_{\mathrm{avg}}
=
\frac{1}{N-1}
\sum_{i=1}^{N-1}
\sqrt{
(x_{i+1}-x_i)^2
+
(y_{i+1}-y_i)^2
},
\end{equation}
and the average control variation is defined as
\begin{equation}
c_{\mathrm{avg}}
=
\frac{1}{N-1}
\sum_{i=1}^{N-1}
\left[
(\delta_{i+1}-\delta_i)^2
+
(a_{\mathrm{lon},i+1}-a_{\mathrm{lon},i})^2
\right].
\end{equation}

To balance reward scales, $\ell_{\mathrm{avg}}$ and $c_{\mathrm{avg}}$ are normalized using robust z-score statistics. The corresponding efficiency and smoothness rewards are defined as
\begin{equation}
r_{\mathrm{len}}
=
-\lambda_{\mathrm{len}}\,\bar{\ell}_{\mathrm{avg}},
\quad
r_{\mathrm{ctrl}}
=
-\lambda_{\mathrm{ctrl}}\,\bar{c}_{\mathrm{avg}},
\label{eq:reward_len_ctrl}
\end{equation}
where $\bar{\ell}_{\mathrm{avg}}$ and $\bar{c}_{\mathrm{avg}}$ denote the normalized path-length metric and control variation, respectively, and $\lambda_{\mathrm{len}}$ and $\lambda_{\mathrm{ctrl}}$ determine the relative importance of trajectory efficiency and control smoothness.

\begin{figure}[t]
  \centering  
  \includegraphics[width=0.9\columnwidth]{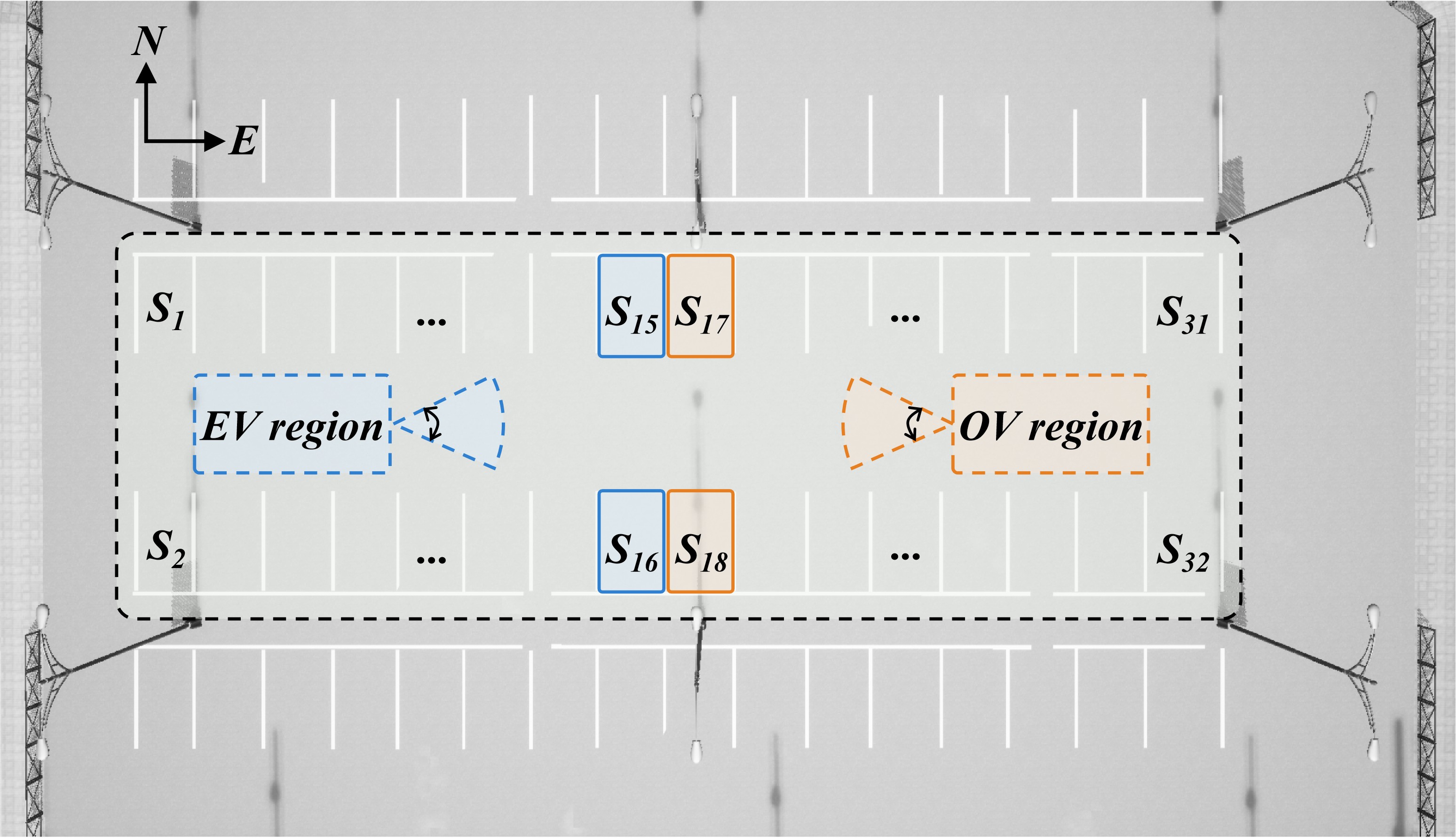}
  \caption{Top view of the parking lot in CARLA.}
  \label{fig:parking-lot}
\end{figure}

\section{EXPERIMENTS}

\subsection{Experiment Setup}

\subsubsection{Implementation Details}

All experiments were conducted on a single NVIDIA A100 GPU with 40\,GB memory. For the proposed model, trainable modules were optimized using AdamW with a mini-batch size of 256. The state encoder was initialized from Behavior Prior Representation pretraining \cite{zang2022behavior} and kept frozen throughout training. The action tokenizer used a vocabulary size of $|\mathcal{Q}|=\text{24}$ and a commitment weight of $\lambda_{\mathrm{com}}=\text{0.25}$. It was trained with a learning rate of $\text{3}\times\text{10}^{-\text{4}}$, and the codebook embeddings were updated by EMA with a decay coefficient of 0.99. The Q-network was then trained with the discrete CQL objective using an initial learning rate of $\text{3}\times\text{10}^{-\text{4}}$ and a cosine decay schedule. The discount factor and conservative regularization weight were set to $\gamma=\text{0.99}$ and $\alpha_{\mathrm{cql}}=\text{1.0}$, respectively.

\subsubsection{Comparison Methods}

We compare the proposed method with representative learning-based baselines under a shared representation setting. Specifically, all methods use the same state encoder, action tokenizer, and downstream LQR tracker, so the comparison focuses on the policy-learning objective applied to the tokenized offline dataset. Behavior Cloning (BC) is included as a supervised baseline that learns to reproduce dataset actions without reward-guided improvement. Soft Actor-Critic (SAC) \cite{haarnoja2018soft} is adapted to the same offline setting as a maximum-entropy actor--critic baseline, providing a reference for standard off-policy RL without explicit conservatism or behavior regularization. We further include SAC-$N$ \cite{an2021uncertainty} and TD3-BC \cite{fujimoto2021minimalist} as representative offline RL baselines. SAC-$N$ introduces pessimism through ensemble-based Q estimation, whereas TD3-BC constrains policy improvement with a BC regularizer.

\subsubsection{Evaluation Metrics}
We evaluate closed-loop parking performance using four mutually exclusive episode-level outcomes. Target Success Rate (TSR) denotes the fraction of episodes in which the vehicle reaches the target pose with a final position error below 1.2$\,\mathrm{m}$ and a final orientation error below 15$^\circ$, without collision or timeout. Target Failure Rate (TFR) denotes episodes that terminate without collision or timeout but do not satisfy the target-pose tolerances. Collision Rate (CR) and Timeout Rate (TR) measure the fractions of episodes terminated by collision and by exceeding the maximum allowed time, respectively. For episodes that do not collide or time out, we further report Average Position Error (APE), Average Orientation Error (AOE), and Average Parking Time (APT), which measure the final Euclidean position error, the final absolute yaw error, and the elapsed time at episode termination, respectively. Building on these success and timing measures, we finally report Success weighted by Completion Time (SCT) \cite{yokoyama2021success}:
\begin{equation}
\mathrm{SCT}
=
\frac{1}{N_e}
\sum_{i=1}^{N_e}
\mathbb{I}_{\mathrm{succ},i}\,
\frac{T_{\mathrm{ref}}}{\max\!\left(T_{\mathrm{ref}},\,T_i\right)},
\end{equation}
where $N_e$ is the number of evaluation episodes, $\mathbb{I}_{\mathrm{succ},i}\in\{0,1\}$ indicates whether episode $i$ succeeds under the TSR criterion, $T_i$ is its parking completion time, and $T_{\mathrm{ref}}$ is a reference parking time. Here, $T_{\mathrm{ref}}$ is simply fixed to 30$\,\mathrm{s}$ across all methods for a fair comparison.

\begin{figure}[t]
  \centering
  \includegraphics[width=0.9\columnwidth]{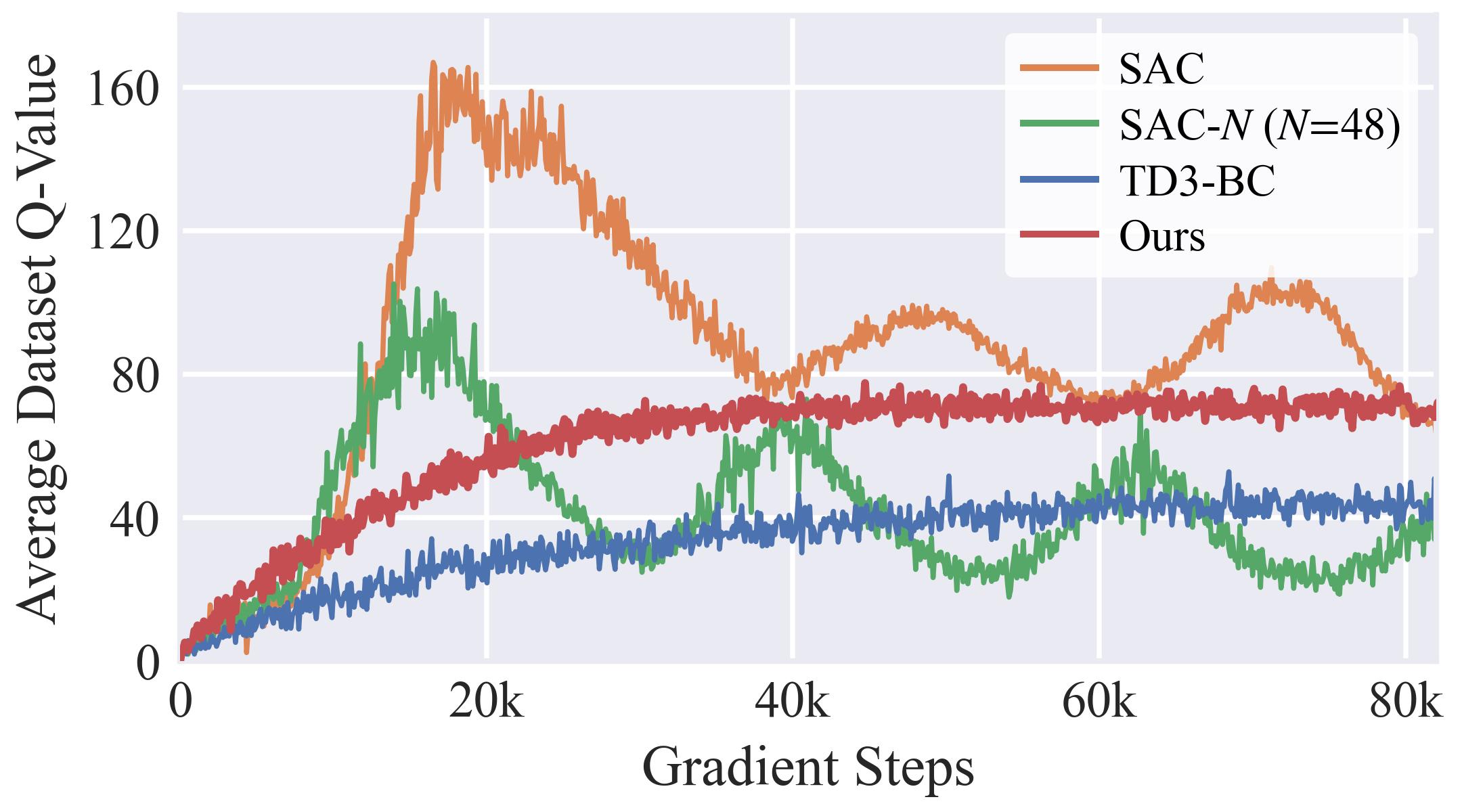}
  \caption{Average dataset Q-value $Q(c_t,q_t)$ during offline training. BC is omitted as it does not learn a value function.}
  \label{fig:q_curve}
\end{figure}

\subsection{Dataset Collection}

The offline parking dataset was collected in CARLA \cite{dosovitskiy2017carla} on the parking lot of the Town04\_Opt map, as illustrated in Fig.~\ref{fig:parking-lot}. We use the two central rows of the parking lot, which contain 32 indexed slots $S_i$, $i=\text{1},\ldots,\text{32}$. The EV is assigned to one of two target slots, $S_\text{15}$ or $S_\text{16}$, and initialized within an $\text{8.0}\,\mathrm{m}\times \text{4.0}\,\mathrm{m}$ region, with its heading sampled from $[-\text{15}^\circ,\text{15}^\circ]$ relative to the eastward direction.

To include both non-interactive and interactive parking behaviors, each episode is sampled with or without an opposite vehicle (OV). When the OV is present, it is initialized on the opposite side of the same aisle and assigned an independent target from $S_\text{17}$ or $S_\text{18}$. Its right-of-way relative to the EV is randomly determined, yielding scenarios in which the EV either proceeds first or yields before entering the aisle.

For each episode, the hierarchical expert pipeline described in Section~\ref{sec:dataset_collection} generates waypoint actions online during data collection. A global Hybrid A* reference is first constructed, followed by local trajectory optimization at 10\,Hz. The optimized waypoint action is perturbed by a rotation angle $\alpha_t\in[-\text{2}^\circ,\text{2}^\circ]$ and then tracked by the LQR controller to advance the simulation. The resulting transition, reward, next state, and terminal indicator are recorded at each decision step. An episode terminates when the EV reaches the target pose, collides with obstacles, or exceeds the time limit. In total, the dataset contains 360 parking episodes.

\begin{table}[t]
\centering
\caption{Ablation results of the action tokenizer.}
\label{tab:ablation}
\begingroup
\small
\setlength{\tabcolsep}{3.0pt}
\begin{tabular}{@{}cccccc@{}}
\toprule
State Cond. & $|\mathcal{Q}|$ & TSR (\%) $\uparrow$ & CR (\%) $\downarrow$ & APE (m) $\downarrow$ & AOE (deg) $\downarrow$ \\
\midrule
             & 24 & 5.56 & 93.75 & 0.99 & 6.30 \\
$\checkmark$ & 24 & 96.53 & 2.78 & 0.71 & 5.26 \\
$\checkmark$ & 36 & 97.92 & 2.08 & 0.73 & 6.63 \\
$\checkmark$ & 48 & 90.28 & 4.86 & 0.81 & 6.94 \\
\bottomrule
\end{tabular}
\endgroup
\end{table}

\begin{table}[t]
\centering
\caption{Generalization results on unseen slots.}
\label{tab:generalization}
\begingroup
\small
\setlength{\tabcolsep}{4.0pt}
\begin{tabular}{@{}ccccc@{}}
\toprule
Slot & TSR (\%) $\uparrow$ & CR (\%) $\downarrow$ & APE (m) $\downarrow$ & AOE (deg) $\downarrow$ \\
\midrule
$S_\text{11}$ & 94.44 & 2.78 & 0.60 & 9.16 \\
$S_\text{12}$ & 88.89 & 2.78 & 0.76 & 6.46 \\
$S_\text{13}$ & 97.22 & 0.00 & 0.95 & 9.52 \\
$S_\text{14}$ & 97.22 & 2.78 & 0.70 & 2.32 \\
Avg.     & 94.44 & 2.08 & 0.75 & 6.88 \\
\bottomrule
\end{tabular}
\endgroup
\end{table}

\subsection{Evaluation Results}

Closed-loop evaluation was conducted in CARLA under the same scenario configuration as dataset collection. Each target slot was evaluated over 36 episodes with different initial poses. The time budget was set to 40$\,\mathrm{s}$ when the OV was present and had priority over the EV, and 20$\,\mathrm{s}$ otherwise. This protocol evaluates the learned policy in both non-interactive and interactive parking scenarios.

\begin{figure*}[t]
  \centering
  \begin{subfigure}{\textwidth}
    \centering
    \includegraphics[width=0.98\textwidth]{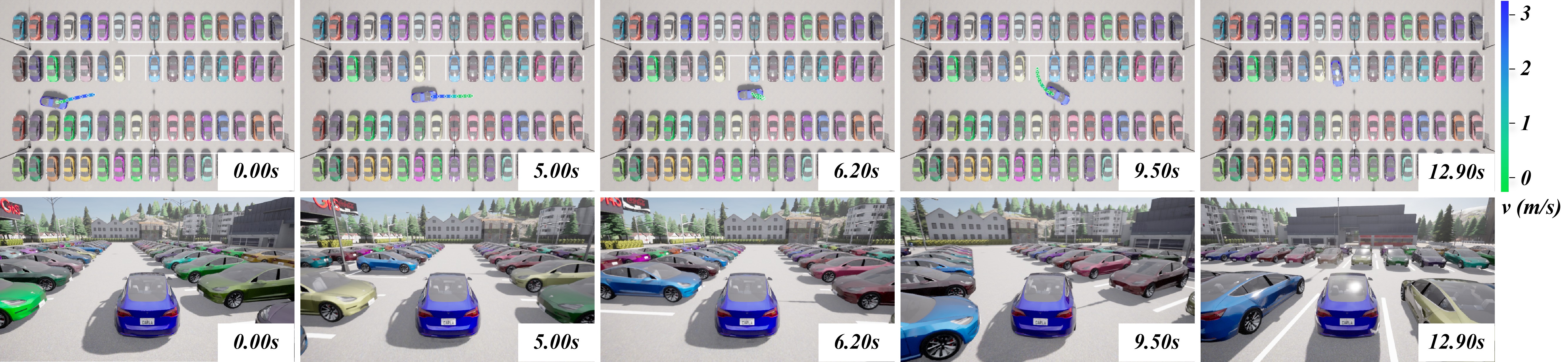}
    \caption{The EV cruises along the aisle and backs into slot $S_{\text{15}}$.}
    \label{fig:demos-a}
  \end{subfigure}
  \begin{subfigure}{\textwidth}
    \centering
    \includegraphics[width=0.98\textwidth]{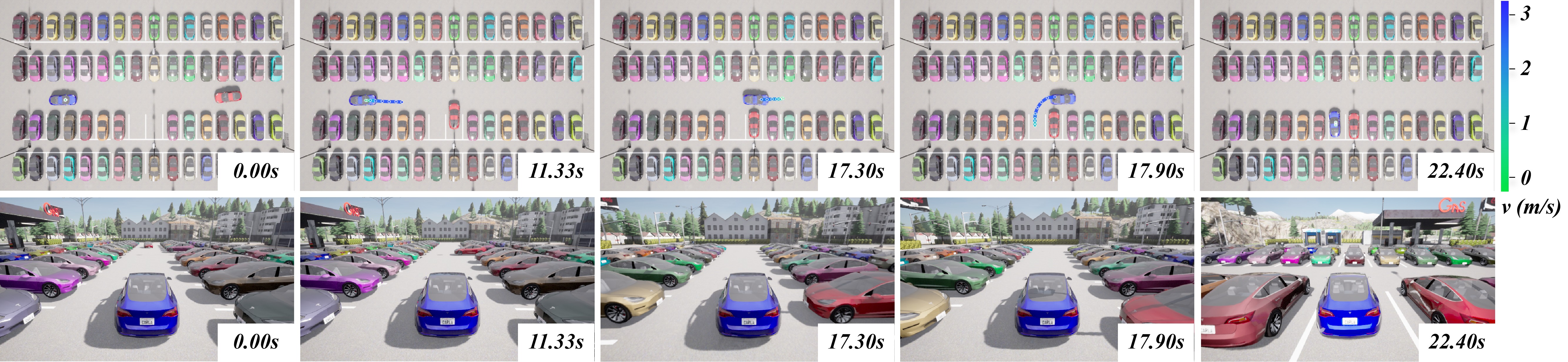}
    \caption{The EV yields to an OV holding right-of-way and then reverses into slot $S_{\text{16}}$.}
    \label{fig:demos-b}
  \end{subfigure}
  \caption{Qualitative demonstrations of closed-loop parking behavior, with waypoints color-coded by speed.}
  \label{fig:demos}
\end{figure*}

\subsubsection{Parking Performance}

Table~\ref{tab:performance} reports the closed-loop performance on the in-distribution target slots $S_\text{15}$ and $S_\text{16}$. The proposed method achieves the highest TSR of 96.53\%, the lowest CR of 2.78\%, and the highest SCT of 95.44\%, with competitive terminal pose errors compared with BC and TD3-BC. BC obtains a nontrivial TSR by imitating waypoint tokens observed in the dataset, but its higher CR suggests that supervised fitting alone cannot reliably distinguish locally similar maneuvers with different long-term safety outcomes. TD3-BC improves over BC by introducing reward-guided policy improvement, yet its BC regularizer still biases the policy toward mixed-quality dataset actions and limits further improvement. SAC fails to achieve successful parking in the offline setting. Without explicit conservatism or behavior regularization, its actor--critic updates can overestimate weakly supported action choices, resulting in ineffective closed-loop behavior dominated by timeouts. Such overestimation is directly reflected in the training curves of Fig.~\ref{fig:q_curve}, which track the average Q-value assigned to dataset action tokens. For SAC, this value rises sharply and thereafter keeps oscillating without converging, a typical manifestation of value divergence induced by bootstrapping from out-of-distribution action tokens. SAC-$N$ partially mitigates this issue through ensemble-based pessimism, but the improvement remains limited when ensemble disagreement does not provide a reliable uncertainty signal for token selection, as its value estimates likewise fail to converge throughout training. By applying conservative value learning directly over discrete waypoint tokens, the proposed method suppresses unsupported candidate actions while exploiting reward information within the offline data distribution, leading to safer and more successful parking behavior. Its value curve accordingly rises smoothly and then remains steady, indicating that the learned value function stays anchored to the action tokens supported by the offline dataset.

\subsubsection{Ablation Analysis}

Beyond the comparison with baseline policies, we further examine the action-token representation in Table~\ref{tab:ablation}. Removing state conditioning reduces the TSR from 96.53\% to 5.56\% and increases the CR from 2.78\% to 93.75\%. This result exposes the representation bottleneck of using a compact codebook, where different parking maneuvers are more likely to be mixed within the same limited token set. With state conditioning, the tokenizer can reuse the compact vocabulary to distinguish maneuver choices under different parking contexts, enabling policy learning in a compact yet discriminative discrete action space. We further vary the vocabulary size to examine sensitivity to codebook granularity. The policy maintains high TSR and low CR under the tested vocabulary sizes, indicating robustness to moderate changes in token resolution. However, finer tokenization also increases terminal pose errors and eventually degrades closed-loop performance. This suggests that excessive codebook granularity can distribute finite offline data over more tokens, weakening token-level value estimation and reducing parking precision.

\subsubsection{Generalization Ability}

To evaluate generalization beyond the slots used for training and in-distribution evaluation, we test the learned policy on unseen slots $S_\text{11}$--$S_\text{14}$, as summarized in Table~\ref{tab:generalization}. The policy achieves an average TSR of 94.44\% and a low CR of 2.08\%, indicating reliable transfer to unseen parking configurations. Although terminal pose errors vary across slots, most episodes are completed successfully without retraining. This suggests that action selection is guided by the current parking context rather than memorized slot-specific trajectory patterns.

\subsection{Qualitative Demonstrations}
Figs.~\ref{fig:demos-a} and~\ref{fig:demos-b} visualize a non-interactive and an interactive closed-loop rollout, respectively. In the non-interactive rollout, the EV is assigned to target slot $S_{\text{15}}$. The policy first guides the EV forward along the aisle, then slows it and adjusts its heading as it approaches the turning point at 6.20\,s. From there, the decoded waypoints form a reverse-in trajectory, and the EV backs into the slot, reaching the target parking pose at 12.90\,s. In the interactive rollout, the EV is assigned to slot $S_{\text{16}}$, while an OV holds priority in the shared aisle. Recognizing that the OV is proceeding through the aisle, the policy chooses to keep the EV stationary and yield, allowing the EV to resume motion only after the conflict clears at 11.33\,s. The EV then executes a forward approach followed by a reverse-in maneuver, completing the parking task at 22.40\,s. These examples illustrate that the learned policy can generate executable waypoint sequences for both direct parking and interaction-aware yielding behaviors.

\section{Conclusions}

This paper presented a waypoint-level offline RL framework for interaction-aware autonomous parking. The proposed framework learns reward-guided waypoint policies entirely from a fixed dataset, thereby eliminating the need for unsafe online exploration. To this end, we constructed a parking dataset from hierarchical expert rollouts spanning both non-interactive and interactive scenarios, discretized the continuous waypoint sequences into state-conditioned action tokens, and performed conservative value learning over these tokens to suppress value overestimation on poorly supported maneuvers. Closed-loop experiments in CARLA showed that the learned policy achieved the highest parking success rate with the fewest collisions among all baselines, and generalized reliably to unseen parking slots. Future work will extend the framework toward denser vehicle interactions and larger-scale parking datasets, and further explore more expressive action representations for complex maneuver generation.

\bibliographystyle{IEEEtran}
\bibliography{refs}

\end{document}